\documentclass[lettersize,journal]{IEEEtran}
\usepackage{amsmath,amsfonts}
\usepackage{algorithmic}
\usepackage{algorithm}
\usepackage{array}
\usepackage[caption=false,font=normalsize,labelfont=sf,textfont=sf]{subfig}
\usepackage{textcomp}
\usepackage{stfloats}
\usepackage{url}
\usepackage{verbatim}
\usepackage{graphicx}
 \usepackage{epstopdf} 
\usepackage{hyperref}
\usepackage{booktabs} 
\usepackage{multirow} 
\usepackage{authblk}
\begin{document}

\title{MSA-GCN:Multiscale Adaptive Graph Convolution Network for Gait Emotion Recognition}

\author{
	\IEEEauthorblockN{Yunfei Yin, Li Jing, Faliang Huang, Guangchao Yang, Zhuowei Wang}
	
\thanks{Yunfei Yin, Li Jing, and Guangchao Yang are with the School of of Computer Science and Technology, Chongqing University, Chongqing, China.}
\thanks{Faliang Huang is with Guangxi Key Lab of Human-machine Interaction and Intelligent Decision, Nanning Normal University, Nanning 530001, PR China.~~Email: {faliang.huang@gmail.com}.}
\thanks{Zhuowei Wang is with the Australian Artificial Intelligence Institute (AAII), Faculty of Engineering and Information Technology, University of Technology Sydney.}
\thanks{Corresponding Author: Faliang Huang.}
}


\maketitle

\begin{abstract}
Gait emotion recognition plays a crucial role in the intelligent system. Most of the existing methods recognize emotions by focusing on local actions over time. However, they ignore that the effective distances of different emotions in the time domain are different, and the local actions during walking are quite similar. Thus, emotions should be represented by global states instead of indirect local actions. To address these issues, a novel \textbf{M}ulti\textbf{S}cale \textbf{A}daptive \textbf{G}raph \textbf{C}onvolution \textbf{N}etwork (MSA-GCN) is presented in this work through constructing dynamic temporal receptive fields and designing multiscale information aggregation to recognize emotions. In our model, a adaptive selective spatial-temporal graph convolution is designed to select the convolution kernel dynamically to obtain the soft spatio-temporal features of different emotions. Moreover, a Cross-Scale mapping Fusion Mechanism (CSFM) is designed to construct an adaptive adjacency matrix to enhance information interaction and reduce redundancy. Compared with previous state-of-the-art methods, the proposed method achieves the best performance on two public datasets, improving the mAP by 2\%. We also conduct extensive ablations studies to show the effectiveness of different components in our methods.
\end{abstract}


\section{Introduction}
\IEEEPARstart{E}{motion} recognition plays an important role in the field of human-computer interaction \cite{5} with applications in various fields, including behavior prediction\cite{1}, video surveillance \cite{2}, character generation \cite{3}, etc. There are many cues of emotion recognition, such as speeches \cite{62,37}, texts \cite{10}, facial expressions \cite{63,7,9}, gestures \cite{8}, and gaits \cite{13}. In the above mentioned cues, gait is defined as an ordered time sequence of joint transformations during a single walking cycle. Compared with other verbal and non-verbal cues, it can be collected from a long distance without human cooperration. And gait is also not easily imitated or deliberately forged \cite{14}. Some psychological works show that individuals in different emotional states may have differences in some gait characteristics, such as arm swing, speed, angle, stride length, etc. \cite{15,16,17}. Therefore, gait emotion recognition is a promising research field.
\begin{figure}[!t]
	\centering
	\includegraphics[scale=0.2]{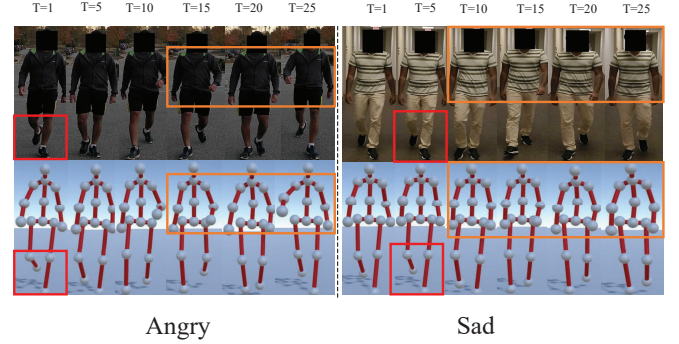}
	\caption{Gait of people with different emotions at different frames. Red box: People with different emotions have similar local actions. Orange box: Some significant discriminative actions of different emotions have different ranges in the time domain.}
	\label{fig1}
\end{figure}

According to the feature extraction schemes, current mainstream gait emotion recognition methods can be roughly divided into two categories: hand-crafted methods and deep learning methods. The hand-crafted method perceives emotions by calculating walking speed, joint amplitude increase, chest flexion, etc. \cite{23}. However, the features obtained by hand-crafted methods usually need to be designed according to the characteristics of the data, and thus, only work well for a certain data set. When the data source changes, the features need to be redesigned, which are very inflexible and inefficient. 

Deep learning based methods extract emotional features through CNN \cite{5,27}, LSTM \cite{13,26} or GCN \cite{28,29}. Because the human skeleton is naturally a graph in non-Euclidean space and 3D joint points are directly modeled by GCN, using GCN for gait emotion recognition is the most commonly used method. However, some experts reveal that GCN based deep learning methods can not model comprehensively enough emotional information \cite{29}, which in turn leads to an overall limitation of their performance in gait emotion recognition tasks. Consequently, how to solve the shortcomings of existing feature representation models is still an active area of research.

To provide a visual illustration, a skeleton diagram of angry and sad people with frames 1, 5, 10, 15, 20, 25 is shown in Figure \ref{fig1}. There is a phenomenon that the two people express different emotions while walking, but the local movements of them are quite similar. For example, from the lower limb movements of the angry person in frame 1 and the sad person in frame 5 marked by the red box, the two people express different emotions, while the movements in the local joints are similar. This means that emotions can not be identified by relying only on local features. However, the existing GCN methods extract spatial information only by modeling local relationships, ignoring the overall state of the gait is the key to identifying emotions, which will increase unnecessary and redundant information. 

In addition, a more prominent and discriminative emotional representation for the anger is the swing amplitude of the shoulders and arms in the last three frames from the yellow box. A more pronounced and discriminative emotional expression for the sad is the forward tilt of the back and the collapse of the spine in the last four frames. It proves that different emotions have different effective ranges in the time domain. But, the existing GCN method will weaken the effectively important information in the time domain by using the same receptive field caused by the same time convolution kernel, which further leads to an overall performance degradation. 


To address the above two challenges of GCN, we propose a \textbf{M}ulti\textbf{S}cale \textbf{A}daptive \textbf{G}raph \textbf{C}onvolution \textbf{N}etwork (MSA-GCN) to obtain more discriminative and robust emotional features to recognize emotion. First, we present an adaptively selected spatio-temporal graph convolution to solve the problem of rigidity in extracting temporal features. Second, we design a new global-local mapping fusion mechanism to obtain discriminative expression of emotion, reduce redundant information and enhance global representation. Specifically, coarse-grained graphs are used to extract overall information. Fine-grained graphs are used to extract local information. Then, the scales features at different scales are fused in a mutually instructive way. The main contributions of this paper are as follows:

\begin{enumerate}
	\item{We propose a novel multiscale adaptive graph convolution network to extract multi-scale emotional features and implement efficient perception of discrete emotions such as happy, angry, sad, and neutral.}
	\item{We design two key components, adaptive selective spatial-temporal graph convolution and cross-scale mapping fusion mechanism, to obtain the soft spatio-temporal features, enhance information interaction, and reduce redundancy.}
	\item{We conduct extensive experiments to show that the proposed MSK-GCN outperforms most state-of-the-art methods for gait emotion perception on two public datasets.}
\end{enumerate}

The rest of this paper is organized as follows. We review previous works related to our proposed model in Section II. Section III describes the proposed model, MSA-GCN, in detail. We present experiment results on two public datasets and conduct detailed discussion in Section IV. Finally, we conclude this paper in Section V.

\section{Related Work}
In this section, we briefly review prior work in classifying perceived emotions from gaits, as well as the related task of skeleton-based action recognition.
\subsection{Skeleton-based action recognition}
Skeleton-based action recognition \cite{18,57} is to identify human movements through 3D key points, such as waving, touching the head, etc. It usually obtains effective spatiotemporal features by modeling the spatial and temporal domains of skeleton to identify different actions. In recent years, action recognition has become a hot research direction in the field of computer vision, which is of great significance to applications such as human-computer interaction and intelligent monitoring.

Deep learning-based action recognition methods initially use RNN, LSTM, and other networks to extract human action features. For example, Wang et al. \cite{45} used a Fourier temporal pyramid to model the relative position of the skeleton. Liu et al. \cite{46} learned the temporal and spatial relationships of skeleton sequences through LSTM and used a gate to remove noise. Although these methods can handle the temporal and spatial information of joint points, they did not fully exploit the natural graph structure of joint points. The graph convolutional network (GCN) that appeared in recent years can make full use of the connection relationship between nodes to model data, which was very suitable for action recognition based on skeleton. Therefore, Yan et al. \cite{18} took the lead in introducing graph convolutional network into skeletal action recognition, and proposed a spatio-temporal graph convolutional network (ST-GCN). ST-GCN represented the skeleton sequence as a spatio-temporal graph and used a graph convolutional layer based on the distance of nodes for feature extraction. On the basis of ST-GCN, Zhang et al. \cite{47} suggested that the bone could also provide information for the network to learn. They proposed an end-to-end learning network that optimizes the topology of the graph and the network, and introduced bone and joint information as a two-stream model to improve accuracy. Considering both spatial and temporal modeling, Liu et al. \cite{49} proposed a unified approach to directly capture complex joint correlations across time and space domains. Zheng et al. \cite{50} considered the self-attention mechanism, which can capture the intrinsic correlation of long-term input. They proposed a convolution-free spatio-temporal Transformer network to extract features. Lei et al. \cite{56} represented a pose-based graph convolutional networks, which use both video data and skeleton data to model.


In the above mentioned literatures, the method of Liu et al. \cite{49} is similar to our method, which also adopts the multi-scale method to extract features. The difference is that our method can refine the fine-grained feature learning by coarse grain instead of simple scale fusion. Zheng et al. \cite{50} used Transformer to model features, which can capture long-term input. However, compared with our method of using GCN for modeling, the input 3D point data needs to be processed in Transformer first, leading to partial information loss.

\begin{figure*}[!t]
	\centering
	\includegraphics[scale=0.235]{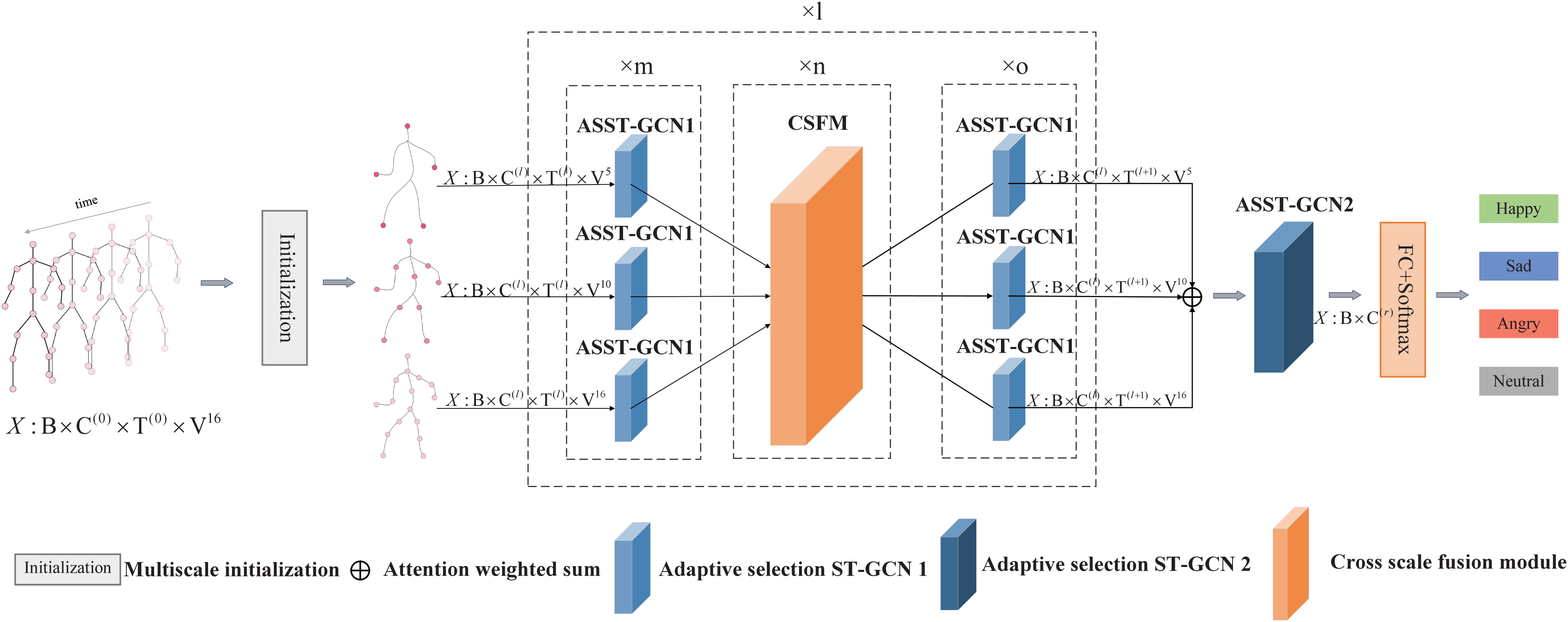}
	\caption{The architecture of MSK-GCN, which mainly uses adaptive selection ST-GCN and cross-scale mapping fusion module to classify emotion.(To better describe the architecture, ASST-GCN in the stacked block was named ASST-GCN1 and ASST-GCN outside the stacked block was named ASST-GCN2. The two modules are essentially the same.)}
	\label{fig2}
\end{figure*}
\subsection{Emotion recognition}
Emotions are a series of complex physiological and psychological reactions produced by people's cognition and perception of the outside world \cite{57}. At present, many researchers have been working on how to make computers capable of observing, understanding and even expressing various emotions like humans \cite{52}. 

For facial expressions, Zhang et al. \cite{63} propoesd a region-based multiscale network that learns features for the local affective region as well as the broad context for emotion image recognition. Mollahosseini et al. \cite{33} combined AlexNet \cite{34} and GoogleNet \cite{35} to recognize facial expressions.For language and tone, Lee et al. \cite{38} used bi-directional long short-term memory (Bi-LSTM) to extract high-level temporal dynamic emotion features. Nie et al. \cite{62} considered the correlation of the intra-class and inter-class videos, and proposed a correlation based graph convolutional network for audio-video emotion recognition. Some scholars also consider using multiple modalities to improve the performance of emotion recognition. For example, Zheng et al. \cite{64} proposed a DNNs-based multi-channel weight-sharing autoencoder with Cascade multi-head attention to address the emotional heterogeneity gap in multimodality.

Although the accuracy of these cues has reached a high degree, they are not effective in some situations, such as occlusion and scene noise. Gait can be used as an auxiliary emotional cue to achieve better emotional recognition in a wider range of scenes.

In the early stage of gait emotion recognition, hand-crafted features were mainly used to identify emotion. For example, Li et al. \cite{20,21} used Fourier transform and principal component analysis (PCA) to extract gait features from the 3D coordinates of 14 human body joint points for emotion recognition. Crenn et al. \cite{22} used hand-designed gait features, employed support vector machines, and introduced a cost function to generate neutral actions for emotion recognition. 

Recently, deep learning has achieved great success in the field of computer vision, and a large number of methods use deep learning for gait emotion perception. Randhavane et al. \cite{13} adopted a time sequence-based approach, using an LSTM to extract temporal features, and then combined it with hand-extracted emotional features for classification using a random forest classifier. Bhattacharya et al. \cite{26} utilized gated recursive units (GRUs) to extract features from joint coordinates at a single time step and perform temporal analysis on them to identify emotions. Narayanan et al. \cite{5} adopted an image-based method to encode skeleton sequences, convert 3D joint point data into images, and then used convolutional neural networks (CNN) to extract features related to emotions in images to identify emotions. Bhattacharya et al. \cite{19} adopted a graph-based method to exploit the characteristic that skeletons are naturally graphs in non-Euclidean space. They expressed the relationship of joints through a spatio-temporal graph convolutional network (ST-GCN) \cite{18}, and extracted spatiotemporal features to classify. Sheng et al. \cite{28} adopted an attention module and proposed an attention-enhanced temporal convolutional network (AT-GCN) to capture discriminative features in spatial dependencies and temporal dynamics for sentiment classification. Zhuang et al. \cite{29} also adopted a graph-based approach to extract global features by constructing global connections and proposed shrinking blocks to reduce noise to improve classification performance. Similarity, Sheng et al. \cite{54} also considered the noise of data, and proposed a multi-task Learning with denoising transformer network to identify emotions. Bhatia et al. \cite{55} presented a light architecture based on LSTM to reduce inference time for gait samples.

Different from the above-mentioned methods which only model emotion features from a single-scale, the method proposed in this paper is based on multi-scale to obtain richer spatial and semantic information. It is worth noting that our model MSA-GCN is similar to G-GCSN \cite{29} proposed by Zhuang et al., which tries to learn better emotion representation by finding global information. However, there is a distinct difference between the two methods. G-GCSN searches for global information by artificially adding edges from the center node to other nodes when constructing the skeleton graph, limiting the expression ability of the network. Our method is more flexible and adaptable because MSA-GCN obtains global information from the constructed coarse-grained maps in a multi-scale manner.

\begin{figure}[!t]
	\centering
	\includegraphics[scale=0.22]{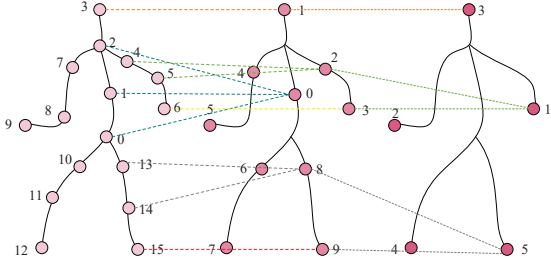}
	\caption{Multi-scale initialization mapping relationship. The three scales consist of 16, 10, and 5 nodes, respectively. Lines of the same color indicate the mapping between nodes of two scale.}
	\label{fig33}
\end{figure}
\section{The proposed MSA-GCN}
In this section, MSA-GCN is introduced in detail. In Section A, a brief overview of our framework is given. In Section B, the input initialization operation is given. In Sections C and D, the two proposed modules, adaptively selected spatio-temporal graph convolution and cross-scale mapping fusion mechanism, are discussed in detail. Finally, the data flow and the loss function of the model are introduced in Section E.

\subsection{Framework}
The framework of our approach is illustrated in Figure \ref{fig2}. On a high level, it contains a stack of $l$ adaptively selected spatio-temporal graph convolution (ASST-GCN) blocks and cross-scale mapping fusion module (CSFM) bolcks to extract features from skeleton sequences after multi-scale initialization, followed by a fully connected layer and a softmax classifier. ASST-GCN and CSFM are cross-deployed to simultaneously capture complex regional spatio-temporal correlations as well as changing spatial and temporal dependencies. First, skeleton maps of different scales are constructed by initializing 3D skeleton sequences. Second, the ASST-GCN1 module (inside the stacked block) is used to extract spatial and temporal features at different scales. Then, the features of different scales are interacted, and fused through the CSFM module, so that the information of one scale can guide the information of another scale. Next, the feature maps of different scales are fused into one scale through the attention mechanism, and then the ASST-GCN2 module (outside the stacked block) is used for overall spatiotemporal feature extraction. Finally, through the fully connected layer and Softmax to obtain classification results: Happy, sad, angry, or neutral.

\subsection{Multiscale initialization}
The multi-scale initialization changes the single scale of the input data to produce coarse-grained scale maps. As shown in Figure \ref{fig33}, the close joints in the fine-grained graph are merged as a joint of the coarse-grained graph, so that the coarse-grained graph can represent semantic information.

The multiple nodes in the fine-grained graph are averaged to represent a node in the coarse-grained scale graph by using 2D average pooling, and then concatenate operation is used to obtain the whole graph representation of the coarse-grained graph.
The formula of multi-scale initialization is expressed as follows:
\begin{equation}
	\label{eq331}
	V_{ck} = pooling(V_{f1} + V_{f2} + \cdots + V_{fh}), k \le h
\end{equation}
\begin{equation}
	\label{eq332}
	Graph_c = concate(V_{c1}, V_{c2}, \cdots , V_{ck})
\end{equation}
where $V_{ck}$ represents the joint information of $k$ nodes in the coarse-grained graph, $V_{fh}$ represents the joint information of $h$ nodes in the fine-grained graph, $Graph_c$ represents the physical skeleton of the coarse-grained graph.

\subsection{Adaptively selected spatio-temporal graph convolution}
ASST-GCN performs spatio-temporal feature extraction from different scales. Considering the extremely complex dynamic characteristics of video data in the temporal dimension, gaits with different emotions present different motion patterns in the temporal dimension. This means the emotion-related features may lack sufficient expressive ability if a convolution kernel is shared for all video data. 

The method of multiple convolution kernels and adaptively select for each person is proposed to aggregate the temporal information. Since each sample has a different receptive field, ASST-GCN can more effectively extract temporal information of different emotions.

\begin{figure}[!t]
	\centering
	\includegraphics[scale=0.22]{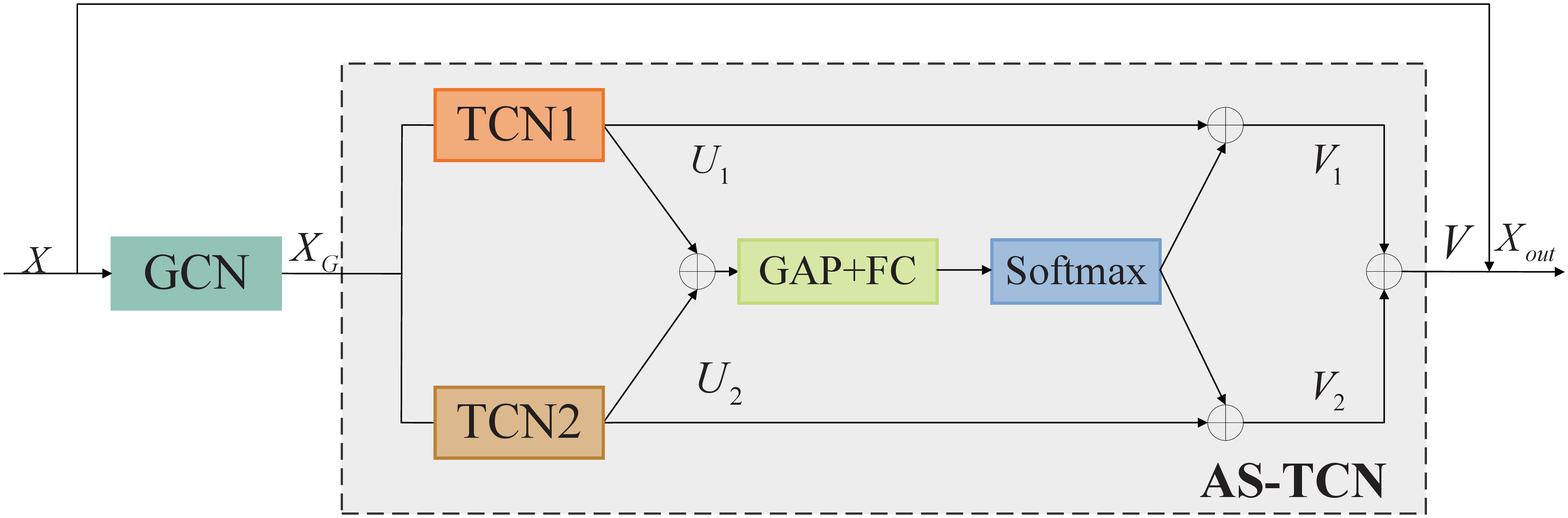}
	\caption{Adaptive selection of spatial temporal graph convolution modules. ASST-GCN adopts a series structure to extract spatio-temporal features. First, the spatial information is extracted through GCN, and then the temporal information is extracted through the AS-TCN structure in the gray box. TCN1 and TCN2 with different convolution kernels are weighted by global average pooling and softmax to obtain adaptive receptive fields.}
	\label{fig3}
\end{figure}
ASST-GCN consists of two components, graph convolutional neural network (GCN) and \textbf{A}ptively \textbf{S}elected \textbf{T}emporal \textbf{C}onvolutional \textbf{N}eural \textbf{N}etwork (AS-TCN), as shown in Figure~\ref{fig3}. GCN is a traditional graph convolution operation, which extracts information in the spatial domain. AS-TCN constructs different receptive fields and extracts features at different levels using multiple convolution kernels. The features are extracted from different convolution kernels through global maximum pooling, fully connected layers, and weighted summation to achieve adaptive selection and fusion of features in the time domain. Finally, the residual connection is added to accelerate convergence and alleviate over-smoothing of graph convolution.

In Figure \ref{fig3}, the light green square represents the graph convolutional neural network, orange and earth-colored squares represent temporal convolution of two different convolution kernels, GAP represents global average pooling, and FC represents a fully connected layer.
The outputs of ASST-GCN are given by Eq. \ref{eq2} - \ref{eq7}.
\begin{equation}
	\label{eq2}
	U_1=TCN_1(ReLu(BN(x_G)))
\end{equation}
\begin{equation}
	\label{eq3}
	U_2=TCN_2(ReLu(BN(x_G)))
\end{equation}
\begin{equation}
	\label{eq4}
	V_1=Softmax(FC(GAP(U_1 \oplus U_2)))) \oplus U_1
\end{equation}
\begin{equation}
	\label{eq5}
	V_2=Softmax(FC(GAP(U_1 \oplus U_2)))) \oplus U_2
\end{equation}
\begin{equation}
	\label{eq6}
	V=V_1\oplus V_2
\end{equation}
\begin{equation}
	\label{eq7}
	X_{out}=V+X
\end{equation}
In equations \ref{eq2} and \ref{eq3}, $x_G$ represents the features after GCN. $U_1$ and $U_2$ represent the output after TCN of different convolution kernels. In equations \ref{eq4} and \ref{eq5}, $V_1$ and $V_2$ represent the results of time convolution after information fusion. In equation \ref{eq6}, $V$ represents the result of AS-TCN. In equation \ref{eq7}, $X_{out}$ represents the output of ASST-GCN.

\subsection{Cross-scale mapping fusion mechanism}
Information at one scale can guide information at another scale \cite{30}. On the human gait represented by the skeleton, the coarse-grained arm information guides the fine-grained hand and elbow information. A cross-scale mapping fusion mechanism is proposed to make the cross-scale information diffusion on different scale graphs, so that information of different scales can interact in the process of feature extraction. CSFM can transform features from one scale to another by building an adjacency matrix between the two scales, i.e. the correspondence between different joints in different graphs. For example, 2 node in second graph can be mapped to 4 and 5 node in first graph in Figure \ref{fig33}.

By constructing an adjacency matrix between two scales, cross-scale fusion can achieve information interaction at different scales and reduce unnecessary redundancy caused by information fusion. The description of CSFM (Figure \ref{fig4}) is as follows.

For each scale, after get the output of ASST-GCN, locally spatially enhanced features are obtained through the attention mechanism to more effectively express the features of different emotions. Then, inspired by the FM model in the recommender system \cite{51}, the inner product of the enhanced features in different scale maps is performed and softmax is used to obtain the correspondence between nodes of different scales, which is the adjacency matrix. Finally, the graph convolution operation is performed to obtain features that fuse other scale information by the adjacency matrix.

Orange and yellow cubes represent features at two scales, green blocks represent spatial attention, orange blocks represent embedding operations, gray blocks represent multilayer perceptron, light green blocks represent ordinary graph convolution operations, and blue cubes represent fusion. Later scale features.

Taking the scale of 16 joints and the scale of 10 joints as an example, we first define the feature maps of the scale of 16 joints and the scale of 10 joints as $X_A$,$X_B$. Then the correspondence between nodes of scales is the adjacency matrix $A_{X_A,X_B}$ and the 16-node feature $\overline{X_{out}}$after scale fusion is: 
\begin{equation}
	\label{eq8}
	\widetilde{X_A}=g_1[embedding(f_1(X_A))]
\end{equation}
\begin{equation}
	\label{eq9}
	\widetilde{X_B}=g_2[embedding(f_2(X_B))]
\end{equation}
\begin{equation}
	\label{eq10}
	A_{X_A,X_B}=softmax(\widetilde{X_B}^\mathrm{T} \cdot \widetilde{X_B}) \in [0,1]
\end{equation}
\begin{equation}
	\label{eq11}
	\widetilde{X_{out}}=GCN(A_{X_A,X_B},X_A)
\end{equation}
$f_1$,$f_2$ represents spatial attention, $g_1$,$g_2$ represents multilayer perceptrons (MLPs), $(\widetilde{X_B}^\mathrm{T} \cdot \widetilde{X_B})$ represents the dot product of $\widetilde{X_B}^\mathrm{T}$ and $\widetilde{X_B}$, $GCN(A_{X_A,X_B},X_A)$indicates that the fused features are obtained by performing graph convolution operations on the scale of 16 nodes through the adjacency matrix $A_{X_A,X_B}$.

\subsection{Data flow and Loss function}
The input of the model consists of $X:B \times C^{(0)} \times T \times V^{16}$. $B$:Batch size, $C^{(0)}$:number of channels in layer 0, $T$:time, $V^{16}$:number of joints is 16 (depended by the dataset). After initialization, we get three branches corresponding to three scales, which only the number of joints $V$ is different. and through the $l+1$-th layer of ASST-GCN and CSFM, the data is  $X:B \times C^{(l+1)} \times T^{(l+1)} \times V^{16}$. Finally, the data is $X:B \times 4$ through the FC layer and Softmax.

Through the whole process, the output of MSA-GCN can be given by Eq. \ref{eq1}:
\begin{equation}
	\label{eq1}
	H(x)=f[att\sum_{X=1}^{X=n}g(f(x))]
\end{equation}
where $H(x)$ represents the output of MSA-GCN,$f(x)$ represents the ASST-GCN, $g(x)$ represents the CSFM, $att\sum_{X=1}^{X=n}$indicates that adaptive attention fusion for feature map of the 1st to the n-th scale.

With the output of MSA-GCN $H(x)$, defined in Eq. \ref{eq1}, FC layer and softmax classifer is used to estimate sentiment distribution of the input $x$, as shown in the following:

\begin{equation}
	\label{eq12}
	p_{i,k} = softmax(fc(H(x)))
\end{equation}
where $p_{i,k}$ denotes  the probability that the $i$-th sample is predicted to be the $k$-th label, and $i = {0,1,...,N}$, $k = {0,1,...,K}$.

For model training, the cross-entropy between the predicted class probabilities and the ground-truth labels is utilized as the loss function of our model.
\begin{equation}
	\label{eq13}
	L = -\frac{1}{N}\sum_{i=0}^{N-1}\sum_{k=0}^{K-1}q_{i,k}\log{p_{i,k}}
\end{equation}
where $q$ and $p$ denotes ground-truth and predicted sentiment
distribution, $K$ the sentiment label set, $N$ the number of samples.

\section{Experiments and Analysis}
In Section A, the datasets and evaluation criteria used in the experiments are given. In Section B, the implementation details and training details of our model are listed. In Section C, the comparison and analysis results with state-of-the-art methods for gait emotion recognition and skeleton-based action recognition are presented. In Section D, the results of ablation experiments on each part are listed. In Section E, the visual analysis is given.
\subsection{Datasets and Evaluation Criterion}
We adopt two public datasets to validate the effectiveness of the method. The first dataset is the Emotion-Gait-16 dataset \cite{42}. It has 16 joints and 240 frame, which consists of 2177 ground-truth gait sequences annotated as happy, sad, angry, and neutral. The second dataset is the Emotion-Gait-21 dataset \cite{26}. The gait is defined as a skeleton with 21 joints, and the step of the gait sequence is 48. It consists of 1835 ground-truth gait sequences with emotional labels provided by 10 annotators, labeled happy, sad, angry, and neutral. 
\begin{table*}
	\begin{center}
		\caption{Comparison with state-of-the-art gait emotion recognition on Emotion-Gait-16 dataset}
		\label{tab1}
		\begin{tabular}{ccccccccc}
			\toprule 
			Method & Happy & Sad & Angry & Neutral & mAP & Precision & Recall & F1\\
			\midrule 
			LSTM(VAnilla)\cite{13} & 0.5528 & 0.7837 & 0.8558 & 0.9135 & 0.7765 & 0.5529 & 0.5431 & 0.5480\\
			TEW\cite{26} & 0.9327 & 0.8507 & 0.9187 & 0.9402 & 0.9106 & 0.7310 & 0.6893 & 0.7095\\
			ProxEmo\cite{5} & 0.8173 & 0.8077 & 0.8462 & 0.8990 & 0.8426 & 0.5749 & 0.5683 & 0.5716\\
			STEP\cite{19} & 0.9375 & 0.8798 & 0.8990 & 0.9183 & 0.9087 & 0.7257 & 0.7234 & 0.7246\\
			G-GCSN\cite{29} & 0.9183 & \pmb{0.9135} & 0.9038 & 0.9087 & 0.9111 & 0.7110 & 0.6789 & 0.6946\\
			MSK-GCN (ours) & \pmb{0.9567} & 0.9038 & \pmb{0.9231} & \pmb{0.9567} & \pmb{0.9351} & \pmb{0.8048} & \pmb{0.8201} & \pmb{0.8124}\\
			\bottomrule 
		\end{tabular}
	\end{center}
\end{table*}

\begin{table*}
	\begin{center}
		\caption{Comparison with state-of-the-art action recognition methods on Emotion-Gait-16 dataset}
		\label{tab2}
		\begin{tabular}{ccccccccc}
			\toprule 
			Method & Happy & Sad & Angry & Neutral & mAP & Precision & Recall & F1\\
			\midrule 
			ST-GCN\cite{18} & 0.9327 & 0.8894 & 0.8942 & 0.9375 & 0.9135 & 0.7520 & 0.7030 & 0.7267\\
			2AS-GCN\cite{47} & 0.9519 & \pmb{0.9375} & 0.9183 & 0.9231 & 0.9327 & 0.7850 & 0.7783 & 0.7816\\
			MS-G3D\cite{49} & 0.9375 & 0.8942 & 0.8798 & \pmb{0.9615} & 0.9183 & 0.7780 & 0.7973 & 0.7875\\
			PoseFormer\cite{50} & 0.5529 & 0.7837 & 0.8606 & 0.9087 & 0.7765 & 0.5529 & 0.5431 & 0.5480\\
			MSK-GCN (ours) & \pmb{0.9567} & 0.9038 & \pmb{0.9231} & 0.9567 & \pmb{0.9351} & \pmb{0.8048} & \pmb{0.8201} & \pmb{0.8124}\\
			\bottomrule 
		\end{tabular}
	\end{center}
\end{table*}

For the evaluation criterion, we quantitatively evaluate our experimental results by the classification accuracy, precision, recall and F1-measure given by formulas \ref{eq12} - \ref{eq15}:
\begin{equation}
	\label{eq12}
	Accuracy=\frac{TP+TN}{TD},
\end{equation}
\begin{equation}
	\label{eq13}
	Precision=\frac{TP}{TP+FP},
\end{equation}
\begin{equation}
	\label{eq14}
	Recall=\frac{TP}{TP+FN},
\end{equation}
\begin{equation}
	\label{eq15}
	F1-Measure=\frac{2\times Precision\times Recall}{Precision+Recall}
\end{equation}
In formulas \ref{eq12} - \ref{eq15}, $TP$,$FP$,$TN$,$FN$ represent the number of true positive, false positive, true negative, and false negative of the four emotions, and TD represents the total data number.

\subsection{Experimental Setup}
All experiments are implemented on an NVIDIA Tesla P40 with the Pytorch framework. 8:1:1 split is used for the training set, validation set, and test set, and the loss function adopts the cross-entropy loss function. The main hyperparameters include the network and training phase. 
For network, when l=1, 2, m, n=1. When l=3, 4, m=1, n=0, 5 and 9 are selected for the temporal convolution kernels in ASST-GCN, the corresponding dimensions are 32, 64, 128, 256. 
For training, we used a batch size of 16 and the Adam optimizer for 400 epochs with an initial learning rate of 0.001. After 200, 300, and 350 epochs, the learning rate decays to one-tenth of its current value. We also used a momentum of 0.9 and a weight decay of $5\times 10^{-4}$. 
\begin{table*}[h]
	\begin{center}
		\caption{Comparison with state-of-the-art gait emotion recognition on Emotion-Gait-21 dataset}
		\label{tab3}
		\begin{tabular}{ccccccccc}
			\toprule 
			Method & Happy & Sad & Angry & Neutral & mAP & Precision & Recall & F1\\
			\midrule 
			LSTM(VAnilla)\cite{13} & \pmb{0.9716} & 0.8750 & 0.9147 & 0.9318 & 0.9233 & 0.5856 & 0.6374 & 0.6104\\
			TEW\cite{26} & 0.9622 & 0.8785 & 0.9235 & 0.9505 & 0.9286 & 0.7257 & 0.7110 & 0.7183\\
			ProxEmo\cite{5} & 0.9545 & 0.8750 & 0.8977 & 0.9205 & 0.9119 & 0.6458 & 0.6302 & 0.6379\\
			STEP\cite{19} & 0.9620 & 0.8696 & 0.9185 & 0.9348 & 0.9212 & 0.6108 & 0.6234 & 0.6170\\
			G-GCSN\cite{29} & 0.9565 & 0.8641 & 0.9185 & 0.9022 & 0.9103 & 0.7109 & 0.7058 & 0.7083\\
			MSK-GCN (ours) & 0.9565 & \pmb{0.9185} & \pmb{0.9348} & \pmb{0.9511} & \pmb{0.9402} & \pmb{0.8074} & \pmb{0.7726} & \pmb{0.7896}\\
			\bottomrule 
		\end{tabular}
	\end{center}
\end{table*}
\begin{table*}
	\begin{center}
		\caption{Comparison with state-of-the-art action recognition methods on Emotion-Gait-21 dataset}
		\label{tab4}
		\begin{tabular}{ccccccccc}
			\toprule 
			Method & Happy & Sad & Angry & Neutral & mAP & Precision & Recall & F1\\
			\midrule 
			ST-GCN\cite{18} & 0.9511 & 0.8587 & 0.9076 & 0.9348 & 0.9131 & 0.7265 & 0.6943 & 0.7100\\
			2AS-GCN\cite{47} & 0.9453 & 0.8906 & 0.9141 & \pmb{0.9531} & 0.9258 & 0.7772 & 0.6940 & 0.7332\\
			MS-G3D\cite{49} & 0.9565 & 0.9022 & 0.9348 & 0.9457 & 0.9348 & 0.7859 & \pmb{0.7786} & 0.7822\\
			PoseFormer\cite{50} & \pmb{0.9674} & 0.8913 & 0.9185 & 0.9076 & 0.9212 & 0.6833 & 0.6529 & 0.6678\\
			MSK-GCN (ours) & 0.9565 & \pmb{0.9185} & \pmb{0.9348} & 0.9511 & \pmb{0.9402} & \pmb{0.8074} & 0.7726 & \pmb{0.7896}\\
			\bottomrule 
		\end{tabular}
	\end{center}
\end{table*}

\subsection{Analysis of results}
MSA-GCN is compared with several state-of-the-art methods of gait emotion recognition. The same dataset partitioning and evaluation method is used for training.

\noindent\textbf{ Emotion-Gait-16.}\quad 
The results of comparison with state-of-the-art gait emotion recognition are shown in Table~\ref{tab1}. The methods used for comparison include sequence-based methods \cite{13,26}, image-based methods \cite{5}, and graph-based methods \cite{19,29}. It can be observed that our method has the best results on Happy, Angry, Neutral, and average accuracy, improving the accuracy by around 0.02-0.05, and the accuracy gap with the best effect in Sad is only about 0.01, which proves that our method is effective. 
In precision, recall, and F1, our method has a large improvement in precision, recall, and F1, and the improvement is all about 0.1, which means that the classification ability of our method on positive samples is excellent, further proving that the higher accuracy is not due to the imbalanced dataset. 

We also compare our method with action recognition to verify that we focus more on the mapping of gait to a set of emotional labels rather than the mapping of actions as shown in Table~\ref{tab2}. Our method achieves the best results on both Happy and Angry, which is about 0.005 higher than the other best results. Although it does not achieve the best results on Sad and Neutral, the accuracy is also high. Neutral is also only about 0.005 lower than the best method. Our average accuracy is also the highest among all methods, which proves that our method finds more emotion-focused mapping than modeling with action recognition.
In precision, recall, and F1, our method has good results and all reaching above 0.8. Especially compared with 2AS-GCN \cite{47}, although this method has the highest accuracy in the Sad class in Table~\ref{tab2}, and mAP is not much lower than our method. So, the classification ability of our method is better, and it is less affected by imbalanced data than 2AS-GCN. It fully demonstrates the superiority of our method.

\noindent\textbf{ Emotion-Gait-21.}\quad 
The results of comparison with state-of-the-art gait emotion recognition are shown in Table~\ref{tab3}. Our proposed method has the best results on sad, Angry, Neutral, and average accuracy, improving the accuracy by around 0.02-0.03.
Our method has a large improvement in precision, recall and, F1, and the improvement is about 0.1, which means that the classification ability of our method on positive samples is excellent. 
From the Table~\ref{tab3}, the accuracy of LSTM (VAnilla) \cite{13} is higher than our method in happy, but F1 is about 0.17 less than our method. The high accuracy of happy maybe since this method prefers to classify samples as happy, rather than classifying samples labeled as happy.

Table~\ref{tab4} presents the results of comparison of our method with the methods for action recognition.
Our method achieves the best results on both Sad and Angry. While on Happy and Neutral, it does not get the best results, but the accuracy is only about 0.002 lower than the best method on Neutral. Our average accuracy is also the highest among all methods, which proves that our method finds more emotion-focused mapping than modeling with action recognition. Our method has the best results in accuracy, and F1, all reaching above 0.78. Although the recall rate does not reach the highest, it is also 0.006 worse than the highest. The 2AS-GCN \cite{47} method and the PoseFormer \cite{50} method are higher than our method in single-class accuracy but far lower than our values in precision, recall, and F1, This shows that these two methods only have a stronger preference for the sentiment of that category, and tend to classify the samples that are not of this category into this category, resulting in high accuracy of this category, rather than learning the classification ability. Therefore, our method has better classification ability in positive samples, which fully demonstrates the superiority of our method.
\begin{figure}[!t]
	\centering
	\includegraphics[scale=0.5]{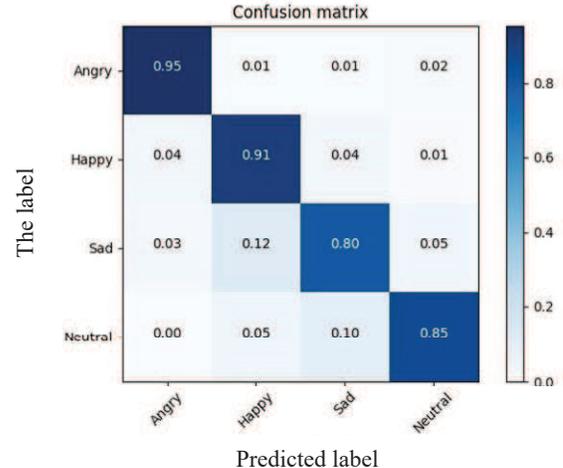}
	\caption{Confusion matrix for MSA-GCN. MSA-GCN has $>$ 80\% accuracy for each class.}
	\label{fig6}
\end{figure}

From Table~\ref{tab1} - \ref{tab4}, mAP is not high on both datasets, which may be because the actions of different emotions in local joints are extremely similar. So, the improved accuracy of our method confirms that it is necessary to consider the global information. The confusion matrix is given to show the discriminatory ability of MSK-GCN in Figure \ref{fig6}. It can be seen that the accuracy of our classifier in each class is greater than 80\%, which means that the classifier can identify each class equally well.

\subsection{Ablation Study}
We perform ablation experiments on our method to highlight the benefit of some crucial elements.

\textbf{Effects of multiple scales.} To verify the proposed multi-scale representation, we employ various scales in MSA-GCN. Besides the three scales in our model, we introduce additional one scale, which represents a body as 3 parts: upper limbs, lower limbs and torso. Table~\ref{tab5} presents the accuracy of various scales. As you can see, two or more scales are most of the time preferable to only using a single scale (1). More scale does not mean better performance. You can see that when combining the four scales (1, 2, 3, 4), the accuracy is not higher than that of a single scale (1). Too high scale will not only bring redundant information and affect the accuracy, but also increase the training time. In addition, the model accuracy reaches the best when three scales (1, 2, 3) are used.

\begin{table}[h]
	\begin{center}
		\caption{Impacts of different scales on the network}
		\label{tab5}
		\begin{tabular}{cccccc}
			\hline
			& \multicolumn{4}{c|}{Node numbers} & Accuracy \\
			\hline
			Scales & 16 & 10 & 5 & 3 &  \\
			\hline
			1 & \checkmark &  &  &  & 0.9159 \\
			1,2 & \checkmark & \checkmark &  &  & 0.9331 \\
			1,3 & \checkmark &  & \checkmark &  & 0.9279 \\
			1,4 & \checkmark &  &  & \checkmark & 0.9135 \\
			1,2,3 & \checkmark & \checkmark & \checkmark &  & \pmb{0.9351} \\
			1,2,4 & \checkmark & \checkmark &  & \checkmark & 0.9159 \\
			1,3,4 & \checkmark &  & \checkmark & \checkmark & 0.9231 \\
			1,2,3,4 & \checkmark & \checkmark & \checkmark & \checkmark & 0.9183 \\
			\hline
		\end{tabular}
\end{center}
\end{table}

\textbf{Effects of CSFM.} The ablation results of CSFM are shown in Table~\ref{tab6}. The baseline is a single-scale graph (16 joints) through ST-GCN networks with channels 32, 64, 128, and 256 without CSFM module and adaptive attention fusion module (ATT). Compared with the backbone network without the CSFM, the model recognition performance with CSFM is significantly improved. And ATT (i.e., spatial attention and channel attention) adopted in the fusion process proves to be useful for emotion recognition.

It can be seen that the CSFM  works best with two layers, which proves that the cross-scale interactive fusion proposed by us is effective. When the level is greater than 2, the accuracy begins to decrease, which may be due to the number of fusion layers being too high, the features tend to be homogenized, and the discriminative emotional features are lost.
\begin{table}[h]
	\begin{center}
		\caption{The effect of the number of scale fusion layers on the accuracy.}
		\label{tab6}
		\begin{tabular}{cccccc}
			\hline
			\multicolumn{5}{c}{Method} &  mAP \\
			\hline
			\multicolumn{5}{c}{MSK-GCN without CSFM, ATT(Baseline)} &  0.9062 \\
			\multicolumn{5}{c}{ATT} & 0.9183 \\
			\hline
			level & 1 & 2 & 3 & 4 &  \\
			\hline

			& \checkmark &  &  &  & 0.9207 \\
			Baseline with CSFMs & \checkmark & \checkmark &  &  & \pmb{0.9255} \\
			& \checkmark & \checkmark & \checkmark &  & 0.8870 \\
			& \checkmark & \checkmark & \checkmark & \checkmark & 0.8029 \\
			\hline
		\end{tabular}
	\end{center}
\end{table}

\begin{figure*}[!t]
	\centering
	\includegraphics[scale=0.15]{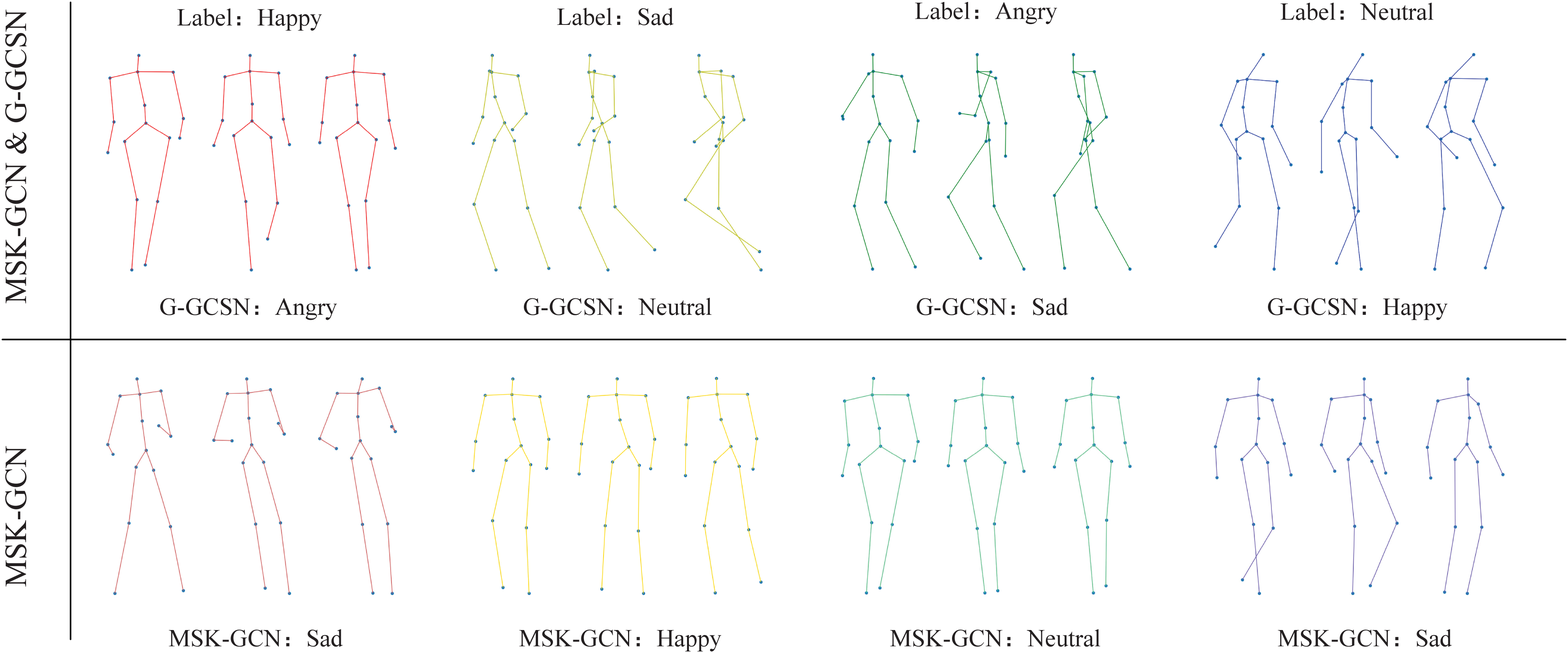}
	\caption{Visual result analysis. The top row shows 4 gaits from the Emotion-Gait dataset where the predicted labels of our network exactly matched the annotated input labels and the predicted labels of G-GCSN did not matched the annotated input labels. The bottom row shows 4 gaits where the predicted labels did not match any of the input labels. Each gait is represented by 3 poses in temporal sequence from left to right. 
	}
	\label{fig7}
\end{figure*}
\textbf{Effects of ASST-GCN.} To verify the effectiveness of ASST-GCN framework, ASST-GCN was compared with traditional ST-GCN \cite{18} (with a single temporal convolution TCN) and ASST-GCN without TCN, as shown in Table~\ref{tab7}. It can be seen from the results without and with TCN that the temporal convolution has a great impact on the accuracy. Compared with the traditional ST-GCN with only a single TCN, our method provides a huge improvement in accuracy. This proves the effectiveness of our proposed framework.
\begin{table}[h]
	\begin{center}
		\caption{Ablation experiments on ASST-GCN}
		\label{tab7}
		\begin{tabular}{cccc}
			\hline
			Method & with one tcn(ST-GCN) & without tcn & ASST-GCN \\
			\hline
			Accuracy & 0.9111 & 0.7764 & \pmb{0.9351} \\
			\hline
		\end{tabular}
\end{center}
\end{table}

\textbf{Effects of SA-TCN.} The ablation results of AS-TCN are shown in Table~\ref{tab8}. The baseline is the ST-GCN network with a single convolution kernel 5 and a multi-scale graph (joints 16, 10, and 5). Compared with the baseline without Adaptively selected temporal graph convolution (AS-TCN), the recognition performance of models with adaptive selection is improved, which means that our proposed AS-TCN is effective. The convolution kernel has the highest performance when selecting 5 and 9. For some convolution kernels with large differences, such as 5 and 75, the effect is not significant. This may be due to the receptive field of the convolution kernel being too large, and invalid information is extracted instead.
\begin{table}[h]
	\begin{center}
		\caption{Impact of kernel size selection on AS-TCN.}
		\label{tab8}
		\begin{tabular}{ccccccc}
			\hline
			\multicolumn{6}{c}{Method} &  Accuracy \\
			\hline
			\multicolumn{6}{c}{MSK-GCN without AS-TCN(Baseline)} &  0.9183 \\
			\hline
			Kernel & 5 & 7 & 9 & 25 & 75 & \\
			& \checkmark & \checkmark &  &  &  & 0.9200 \\
			& \checkmark &  & \checkmark &  &  & \pmb{0.9351} \\
			&  & \checkmark & \checkmark &  &  & 0.9327 \\
			& \checkmark &  &  & \checkmark &  & 0.9279 \\
			& \checkmark &  &  &  & \checkmark & 0.9207 \\
			&  &  & \checkmark & \checkmark &  & 0.9228 \\
			\hline
		\end{tabular}
	\end{center}
\end{table}

\subsection{Visual Analysis}
To demonstrate the robustness and limitation of the proposed method, a visualization of the correct classification of MSA-GCN and the wrong classification of G-GCSN (the best gait emotion recognition method other than ours in the experiment) under the same sample (top) and a visualization of the wrong of MSA-GCN (bottom) are shown in Figure \ref{fig7}. 

Comparing between MSK-GCN and G-GCSN (top), there are fewer samples classified correctly by G-GCSN and wrong by MSK-GCN, so only the samples classified correctly by MSA-GCN and wrongly classified by G-GCSN are listed. The samples given in the upper part have an obvious feature. The collected samples are not facing the camera directly, but have an oblique angle (except for the first sample). Such samples are much more difficult to classify than normal front-facing samples. Because the joint points are more likely to overlap and obscure the important emotional representation of the shoulder. Comparing the sad and angry samples with the same left-facing side, the lower limb movements of the two samples are basically the same, and the upper limbs tend to be the same due to the overlapping joint points of walking, which may be the reason why G-GCSN classifies the angry samples as the label sad. This also proves that we mentioned in the introduction that when the actions of local joints are similar, and if only focusing on the relationship of local joints may confuse the conjecture of the classification results.

For the analysis of MSA-GCN misclassification (bottom), the misclassified samples is selected from the class that is more prone to misclassification, which is more conducive to the analysis of why the classification error occurs. It can be seen that when misclassified, the samples tend to be classified as happy, sad, and neutral. Angry was less likely to be classified, it should be that the limbs swayed more and faster when angry. For the four samples that were misclassified, the reason for the misclassification may be that the gait of these samples expressing emotions is relatively restrained, and the reaction is that the degree of posture collapse and the amplitude of hand swing are small on the skeleton, so few discriminant features are extracted by the network, leading to classification is incorrect.

\section{Conclusion}
In this work, we proposed the multiscale adaptive graph convolution network for gait-based emotion recognition. Considering that the emotion of gait is presented by the overall state and the emotion information of every sample in time domain is different, we obtain emotional mapping through a multiscale adaptive graoh convolution network we proposed. We conducted extensive experiments on two public datasets. As a result, we showed the effectiveness of our module and showed that the model in which our module is implemented achieves state-ofthe-art performance compared with previous state-of-the-art methods on both of them. In the future, we will extend our method to noisy label learning \cite{61} and federated learning \cite{60} with gait-based emotion recognition.


\section*{Acknowledgments}
\IEEEcompsocthanksitem This work was supported by the Natural Science Foundation of China under Grant 61962038, Grant 61962006, and by the Guangxi Bagui Teams for Innovation and Research.

\bibliographystyle{IEEEtran}
\bibliography{JLrefer}


\section{Biography Section}
\begin{IEEEbiography}[{\includegraphics[width=1in,height=1.25in,clip,keepaspectratio]{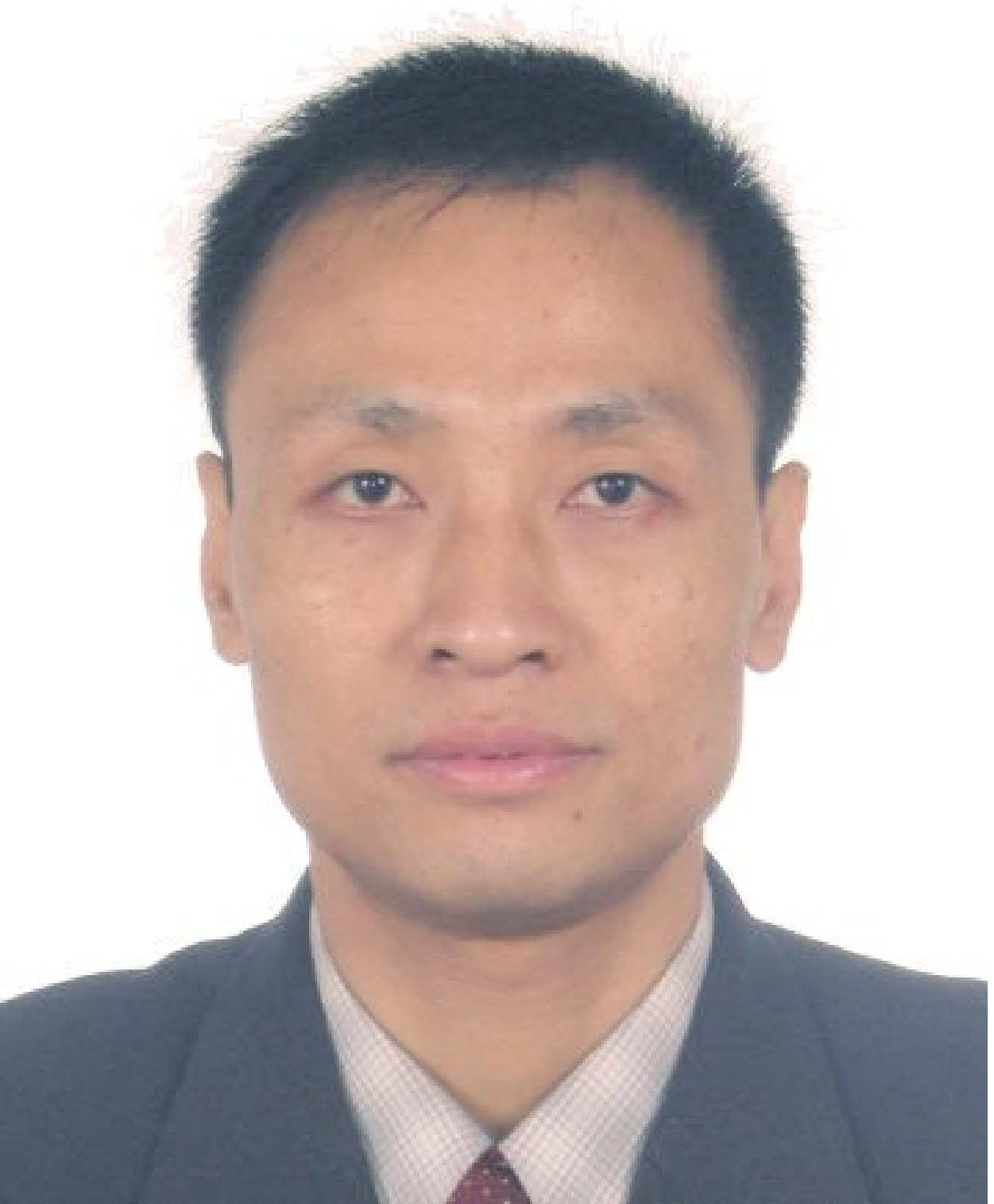}}]{Yunfei Yin}
	received the B.S. degree in Computer Science from Peking University, Beijing City, R.P. China in 2002, the M.S. degree in Computer Engineering from Guangxi Normal University, Guilin City, R.P. China, in 2005 and the Ph.D. degree in Control Science \& Engineering from Beijing University of Aeronautics and Astronautics, Beijing City, R.P. China, in 2010.
	From 2010 to 2013, he was a Research Assistant with the Chongqing University. Since 2013, he has been an Associate Professor with the College of Computer Science, Chongqing University. He is a reviewer for Journal of Software (Chinese), Artificial Intelligence, IEEE International Conference on Data Mining (ICDM), and other journals and conferences. He is mainly engaged in Artificial Intelligence (Data Mining), System Modeling, Computer Simulation and Unmanned Aerial Vehicle researches. In recent years, he has involved in the National Natural Science Foundation of China, International Large Research Foundation, provincial and ministerial level foundations and other projects, a total of 20. He has published more than 40 SCI/EI/ISTP-cited refereed papers.
\end{IEEEbiography}
\begin{IEEEbiography}[{\includegraphics[width=1in,height=1.25in,clip,keepaspectratio]{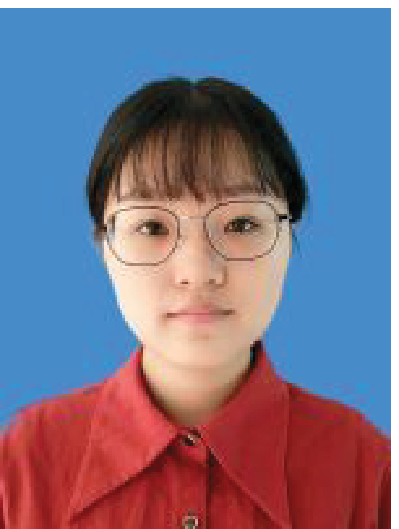}}]{Li Jing}
	is currently working toward the M.S. degree with the School of Computer Science and Technology, Chongqing University, Chongqing, China. Her research interests include emotion recognition and computer vision. Contact her at 202114021068t@cqu.edu.cn
\end{IEEEbiography}
\begin{IEEEbiography}[{\includegraphics[width=1in,height=1.25in,clip,keepaspectratio]{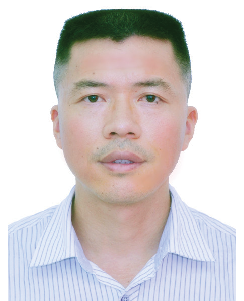}}]{Faliang Huang}
	received the PhD degree in data mining from the South China University of Technology, China.  He is now a professor with the School of Computer and Information Engineering, Nanning Normal University. His research interests include data mining and sentiment analysis.
\end{IEEEbiography}
\begin{IEEEbiography}[{\includegraphics[width=1in,height=1.25in,clip,keepaspectratio]{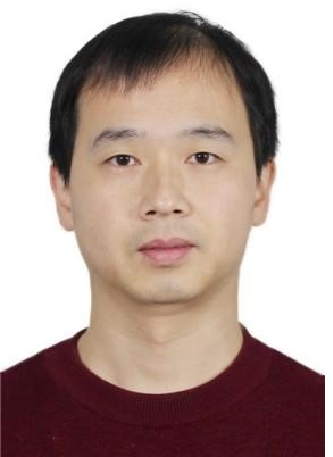}}]{Guangchao Yang}
	received the PHD degree from College of Computer Science, ChongQing University, China in 2017.  He is now an associate professor with the College of Computer Science, ChongQing University. His research interests include pattern recognition and image processing.
\end{IEEEbiography}
\begin{IEEEbiography}[{\includegraphics[width=1in,height=1.25in,clip,keepaspectratio]{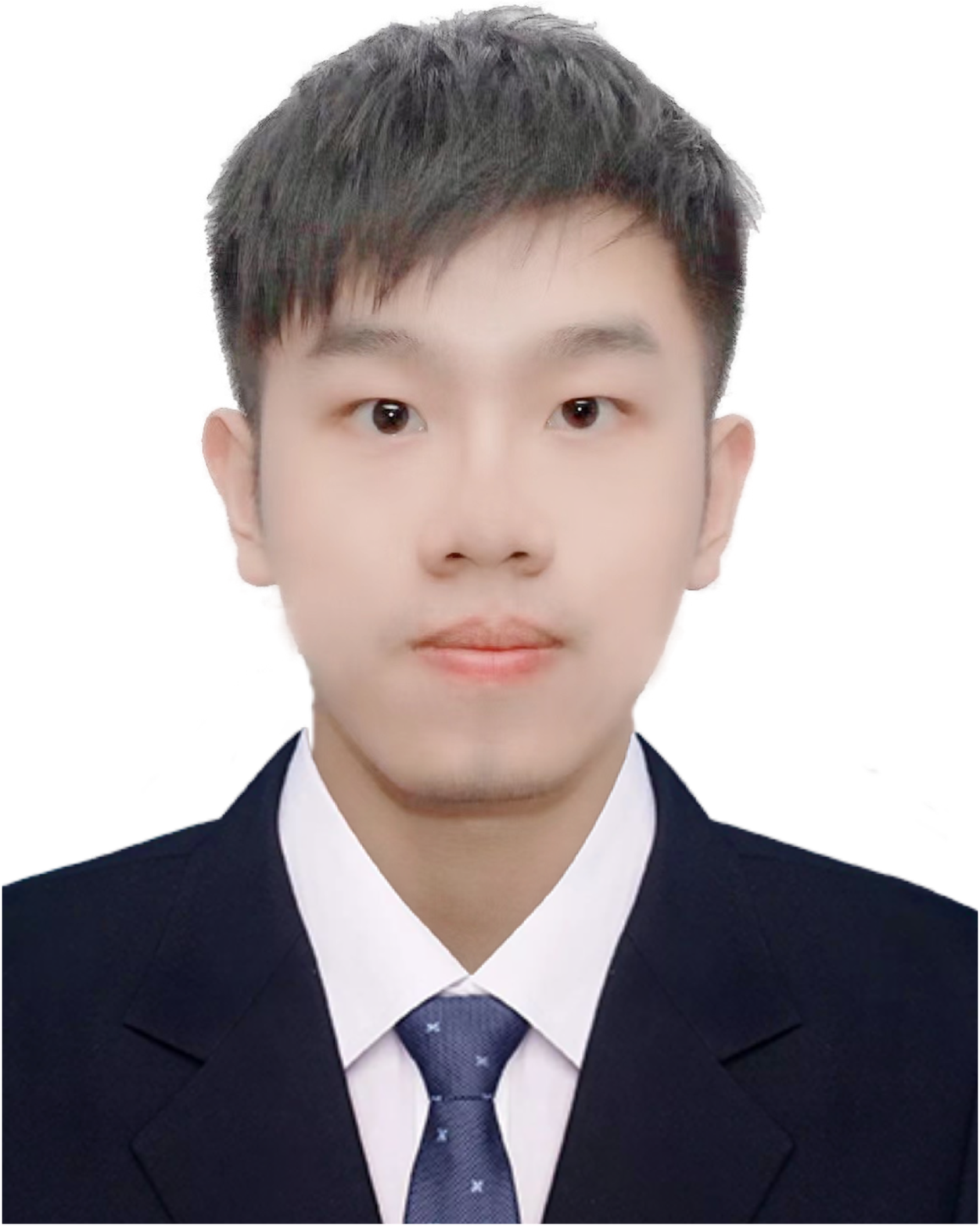}}]{Zhuowei Wang}   received his Ph.D. degree in Computer Science from University of Technology Sydney (2018-2022) and his bachelor's degree from South China University of Technology (2013-2017). His research interest focuses on noisy label learning, federated learning, and weakly supervised learning. He has served as Editorial Manage of Expert Systems with Applications and Technical Committee of BDAI and WSPML. He has served as a member of program committees of CVPR, ICML, ECCV, and KDD.
\end{IEEEbiography}

\vfill

\end{document}